\documentclass[letterpaper, 10 pt, conference]{ieeeconf}  

\IEEEoverridecommandlockouts                              

\usepackage{graphics} 
\usepackage{epsfig} 
\usepackage{mathptmx} 
\usepackage{times} 
\usepackage{amsmath} 
\usepackage{amssymb}  
\usepackage{pgfplots}
\usepackage{import}
\usepackage{xcolor,colortbl}
\usepackage{array}
\newcolumntype{C}[1]{>{\centering\arraybackslash}p{#1}}
\newcolumntype{G}[1]{>{\columncolor{green}\centering\arraybackslash}p{#1}}
\newcolumntype{R}[1]{>{\columncolor{red}\centering\arraybackslash}p{#1}}

\DeclareMathAlphabet{\mathcal}{OMS}{cmsy}{m}{n}
\renewcommand\vec{\mathbf}

\title{\LARGE \bf
Deep Inverse Sensor Models as Priors for evidential Occupancy Mapping
}

\author{Daniel Bauer$^{1}$, Lars Kuhnert$^{1}$ and Lutz Eckstein$^{2}$
\thanks{$^{1}$Daniel Bauer and Lars Kuhnert are with Ford Werke GmbH, Germany
        {\tt\small \{dbauer31, lkuhnert\}@ford.com}}%
\thanks{$^{2}$Lutz Eckstein is with the Institute of Automotive Engineering, RWTH Aachen University, Germany
        {\tt\small office@ika.rwth-aachen.de}}%
}

\begin{document}

\maketitle
\thispagestyle{empty}
\pagestyle{empty}

\begin{abstract}

With the recent boost in autonomous driving, increased attention has been paid on radars as an input for occupancy mapping. Besides their many benefits, the inference of occupied space based on radar detections is notoriously difficult because of the data sparsity and the environment dependent noise (e.g. multipath reflections). Recently, deep learning-based inverse sensor models, from here on called deep ISMs, have been shown to improve over their geometric counterparts in retrieving occupancy information \cite{weston2018probably,sless2019road,bauer2019deep}. Nevertheless, these methods perform a data-driven interpolation which has to be verified later on in the presence of measurements. In this work, we describe a novel approach to integrate deep ISMs together with geometric ISMs into the evidential occupancy mapping framework. Our method leverages both the capabilities of the data-driven approach to initialize cells not yet observable for the geometric model effectively enhancing the perception field and convergence speed, while at the same time use the precision of the geometric ISM to converge to sharp boundaries. We further define a lower limit on the deep ISM estimate's certainty together with analytical proofs of convergence which we use to distinguish cells that are solely allocated by the deep ISM from cells already verified using the geometric approach.

\end{abstract}

\section{INTRODUCTION}
\begin{figure}
	\begin{center}
		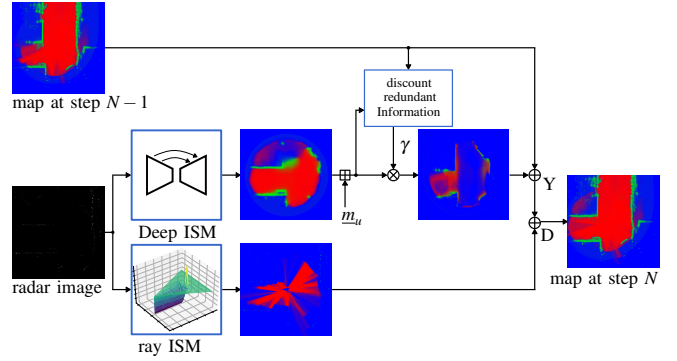
		\caption{\label{fig:architecture}Overview of approach to integrate deep ISM estimates into the evidential occupancy mapping framework, by accounting for redundant information ($\text{rgb} \rightarrow \{\text{free, occupied, unknown}\}$ $\|$ best viewed in color with zoom).}
	\end{center}
\end{figure}

For robotic applications, occupancy maps are a long established form of environment representation \cite{elfes1989using,pagac1998evidential,thrun2005probabilistic}. Because of their robustness towards environmental conditions, radars are heavily relied on to obtain the occupancy information for automotive applications \cite{caesar2019nuscenes}. However, radar data often comes in the form of sparse, noisy detections, based on which it is non-trivial to infer the occupancy state \cite{richards2010principles}. Methods to infer the environment based on the measurement are referred to as inverse sensor models (ISMs) \cite{thrun2005probabilistic} which can be divided into the categories of data-driven and geometric approaches. The geometric ISMs use the sensor's measurement principle to define a geometric correlation between the sensor position and the detection. On the other hand, data-driven models define a machine learning model and use the measurements themselves to tune the ISM. In recent years, many deep learning approaches, from here on referred to as deep ISMs, have been proposed to handle the sparseness of radar data \cite{weston2018probably,sless2019road,bauer2019deep}. For the fusion of predictions over time, it is important to know that geometric ISMs only infer the state of cells which are in a geometric correlation between sensor and detection. Hence, the information provided by two separate measurements is largely independent. In contrast to that, deep ISMs do also infer the occupancy state of areas far away from actual detections. This leads to situations in which two independent measurements can provide mostly the same information about a cell's state. This results in a high amount of redundant information between time steps which, if not accounted for, accumulates and leads to wrong convergence. To deal with this redundancy, we propose the following. First, limit the certainty of the deep ISM estimate to a tuneable threshold $\underline{m}_u$. Afterwards, discount the deep ISM prediction using a discount factor $\gamma$ depending on the amount of redundant information between the current map and the deep ISM. Eventually, combine the non-redundant part of the deep ISM with the map and the geometric ISM's prediction. This procedure is depicted in Fig. \ref{fig:architecture}. 

In summary, our contributions are first, the identification of redundant information as the cause of wrong convergence when integrating deep ISM predictions in the same way as geometric estimates. And secondly, an analytically verified procedure to account for the redundant information in a way that cell states solely allocated by a deep ISM have a lower bound $\underline{m}_u$ on their certainty which makes them distinguishable from pixels verified by the geometric ISM. Additionally, only new information is added in each time step, limiting the accumulation of small evidences.
Using this integration scheme, we are able to leverage both the capabilities of deep ISMs to enhance the perception field and of geometric ISMs to converge to sharp boundaries.

\section{RELATED WORK}
\subsection{Occupancy Mapping}
Occupancy maps are a common representation of a robots environment \cite{carrillo2015autonomous,fleuret2008multicamera,elfes1989using}, where the space around the robot is discretized into a bird's-eye-view grid with each grid cell containing its occupancy information. The two dominant forms of encoding this information is as probabilities, proposed by Elfes \cite{elfes1989using}, or in the Dempster-Shafer evidence framework \cite{pagac1998evidential,dempster1968generalization}. In contrast to the probabilistic formulation, the evidential representation uses an additional degree of freedom which allows to distinguish conflicting information between free and occupied from the state of missing information. This conflict can e.g. be used to indicate dynamic objects \cite{moras2011moving}. On the other hand, the amount of missing information correlates with the degree of exploration which we will exploit in this work to indicate whether pixels have only been allocated using the deep ISM or have also been verified using the geometric ISM.  

\subsection{Combination of independent Evidence}
One central question in the evidential framework is how to combine two evidence masses. Under the assumption of informational independence, several combination rules have been proposed which fall into the categories of associative and non-associative rules \cite{bostrom2008evidential}.   
The two dominant methods of both categories are the associative Dempster's rule \cite{dempster1968generalization} and the non-associative Yager's rule \cite{yager1987dempster}. In our case, the non-associativity of Yager's rule is acceptable, since we always fuse information from the ISMs into the occupancy maps, thus giving us a clear order. In sec. \ref{subsec:choice_of_combination_rule}, we analyse the pros and cons of both methods and explain in which scenario to take which rule.

\subsection{Combination of dependent Evidence}
In this work, we argue that the sources of evidence between two deep ISM predictions are not independent for each pixel. Hence, using the above mentioned combination rules results in unwanted effects, which will be elaborated on in sec. \ref{sec:method}. However, a number of methods have been proposed to handle the combination of dependent evidences. These methods address the problem by either altering the combination rule itself or discounting the masses directly according to the amount of redundant information \cite{su2018research}. Here, we approach the problem by discounting the masses. 

To quantize the amount of redundancy, several measures have been proposed ranging from the relative independence degree \cite{yager2009fusion} to the Pearson \cite{su2014consideration} or Spearman rank correlation coefficient \cite{shi2017research} through to using the mutual information between the two evidence masses \cite{su2018research}. Our approach is closely inspired by \cite{su2018research}. However, in sec. \ref{subsec:choice_of_info_measure_and_discount_fact}, we argue that using the unknown mass instead of the entropy as a measure of information fits better into the evidential occupancy mapping framework.  
  
\subsection{Geometric Inverse Sensor Models}
To model the inverse relationship between the environment and the sensor, one approach is to use geometric information inherent in the measurement principle. In our case, we use lidar and radar sensors which both have a polar measurement scheme. This means that ideally, everything in a line between the sensor and the detection is free and the detection itself is occupied. In case of the lidar, the data is accurate enough to apply the ideal sensor model as shown in Fig. \ref{fig:comparison_isms}. When, however, applied on sparse radar data, the ideal sensor model results in too sparse free space. To account for this, the ideal model can be convolved with e.g. a Gaussian noise \cite{werber2015automotive,bauer2019deep}. This not only enlarges the perceived free space but also accounts for detection inaccuracies. For our radar model, we build upon the work of \cite{werber2015automotive} and adapt it in a way to produce sharper occupied space which is elaborated on in sec. \ref{subsec:ray_ism}.  

\subsection{Data-driven Inverse Sensor Models}
Recently, many deep learning approaches have been proposed to model the inverse radar characteristics. One half of the methods defines the inverse radar characteristic as a probabilistic model \cite{weston2018probably,bauer2019deep} while the other half models it in the evidential framework \cite{sless2019road,bauer2019deep}. Since we choose to use the evidential occupancy mapping framework, we will only consider the second half. In this category, Sless et al. \cite{sless2019road} proposed to train a network using accumulated radar images as input and inverse lidar models as target. The network uses a softmax with cross entropy loss to predict the three categories free, occupied and unknown. Another approach proposed by Wirges et al. \cite{wirges2018evidential} models the inverse lidar model using a multichannel input with detections, transmissions and intensity. As labels, they build occupancy maps and use patches around each vehicle pose to better capture the ground-truth occupancy state. They also use a softmax, however, train on $L_1$ and $L_2$ losses with a specific scaling factor to reduce the influence in presence of label uncertainty. Recently, we have proposed an evidential model \cite{bauer2019deep} which is capable of learning the aleatoric uncertainty and encode it into the unknown mass. We compared it against a model similar to the two mentioned before and could demonstrate some improvements with regards of robustness and segmentation sharpness. Hence, we propose to use this model also in this work. In sec. \ref{subsec:evnet}, we will provide an overview and show some exemplary prediction results.

\section{PRELIMINARIES}
\subsection{Evidential Occupancy Mapping States}
In this section, we present a brief review of the evidential occupancy mapping framework \cite{shafer1992dempster,pagac1998evidential}. For occupancy mapping, the state of a cell can either be free $f$ or occupied $o$. Hence, the set of possible states called frame of discernment $u$, which will be treated as the unknown state, and its power set $2^{u}$ can be defined as 
\begin{align}
	u = \{f, o\} \text{, } s^{u} = \{\emptyset, f, o, u\}
\end{align}
Evidences are represented in the form of mass functions $\vec{m}$ that define the evidence mass for each state as follows 
\begin{align}
	\label{eq:vect_convention}
	\vec{m} &= \begin{bmatrix} m_f, & m_o, & m_u \end{bmatrix}^{\top}\\
	m_\emptyset = 0 \text{, } &m_k \in [0,1] \text{ with } k \in \{f,o,u\}\\
	\label{eq:ev_mass_sum_convention}
	&\sum_{k} m_k = 1
\end{align}
with the evidence mass for free $m_f$, occupied $m_o$ and unknown $m_u$.

\subsection{Evidential combination Rules}
In this work, we only focus on the application of Dempster's and Yager's combination rules. For occupancy mapping, these can be written as follows
\begin{align}
	&\textit{Dempster's Rule}\nonumber\\
	\label{eq:dempsters_rule}
	&\vec{m}_1 \oplus_D \vec{m}_2 = 
	\begin{bmatrix} 
		(m_{f1}m_{f2} + m_{f1}m_{u2} + m_{u1}m_{f2}) / (1-K) \\
		(m_{o1}m_{o2} + m_{o1}m_{u2} + m_{u1}m_{o2}) / (1-K) \\
		m_{u1}m_{u2} / (1-K) 
	\end{bmatrix}\\
	\nonumber\\ 
	&\textit{Yager's Rule}\nonumber\\
	\label{eq:yagers_rule}
	&\vec{m}_1 \oplus_Y \vec{m}_2 = 
	\begin{bmatrix} 
		m_{f1}m_{f2} + m_{f1}m_{u2} + m_{u1}m_{f2} \\
		m_{o1}m_{o2} + m_{o1}m_{u2} + m_{u1}m_{o2} \\
		m_{u1}m_{u2} + K
	\end{bmatrix}
\end{align}
with the conflict $K$ being defined as
\begin{align}
\label{eq:conflict}
	K = m_{f1}m_{o2} + m_{o1}m_{f2}
\end{align}
\subsection{Discounting Evidence}
Given a discounting factor $\gamma \in [0,1]$, the discounting rule can be formulated as follows
\begin{align}
	\label{eq:discount_eq}
	\gamma \otimes \vec{m} = \begin{bmatrix} \gamma m_f, & \gamma m_o, & 1 - \gamma + \gamma m_u \end{bmatrix}^{\top}
\end{align}

\subsection{Subjective Logic Transformation}
Josang \cite{josang2016subjective} proposed a framework which collects evidences $e \in \mathbb{R}_{>0}$ for all but the unknown class which are not equal to evidence masses. These evidences can be used to either define a probabilistic Dirichlet or an evidential mass function. He also includes transformations between the two representations. The formulas relevant for occupancy mapping are summarized below
\begin{align}
	\text{Evidence Vector}&\nonumber\\
	\label{eq:josang_evidence}
	\vec{e} &= \begin{bmatrix}e_f, & e_o\end{bmatrix}^{\top}\\
	S &= 2 + e_f + e_o\\
	\text{Probabilistic View}&\nonumber\\
	\vec{\alpha} &= \vec{e} + I\\
	\vec{p} &= \begin{bmatrix}p_f \\ p_o\end{bmatrix} = \mathbb{E}[Dir(\vec{\alpha})] = \dfrac{\vec{e} + I}{S}\\
	\label{eq:pignistic_prob}
	p_o &= \dfrac{m_u}{2} + m_o = \dfrac{1 - m_f + m_o}{2}\\
	\text{Evidential View}&\nonumber\\
	\label{eq:josang_evidential_view}
	\vec{m} &= \begin{bmatrix}e_f/S, & e_o/S, & 2/S \end{bmatrix}^{\top}
\end{align}

\subsection{Lower Limit on Unknown Mass}
We use the operator $\boxplus$ to ensure that a mass function $\vec{m}$ has at least $\underline{m_u}$ unknown mass assigned as follows  
\begin{align}
\label{eq:lower_un_mass_limit}
	\vec{m} \boxplus \underline{m_u} &= 
	\begin{bmatrix} 
	(1 - \alpha)m_f \\  
	(1 - \alpha)m_o \\ 
	m_u + \Delta m_u
	\end{bmatrix}\\
	\text{with }\Delta m_u &= \max(0;\underline{m_u} - m_u) \text{, } \alpha = \dfrac{\Delta m_u}{m_f+m_o}
\end{align}

\section{METHOD}
\label{sec:method}
\subsection{Choice of Combination Rule}
\label{subsec:choice_of_combination_rule}
To chose an adequate combination rule, we first have to examine the properties of Dempster's and Yager's rules. Dempster's rule is defined in a way that the fused unknown mass can only be lesser or equal to the separate unknown masses 
\begin{align}
\max(m_{u1}, m_{u2}) &\geq m_{u12}
\end{align}
Since Dempster's rule is associative, we only need to show this property with respect to one of the in-going masses
\begin{align}
m_{u1} &\geq m_{u12} = m_{u1}\dfrac{m_{u2}}{1 - K} & &| \text{with \eqref{eq:dempsters_rule}}\\
\Leftrightarrow 1 &\geq \dfrac{1-m_{f2}-m_{o2}}{1-m_{f1}m_{o2} - m_{o1}m_{f2}} & &| \text{with \eqref{eq:conflict}, \eqref{eq:ev_mass_sum_convention}}\\
\Leftrightarrow 0 &\geq m_{f2}\underbrace{(m_{o1}-1)}_{\leq 0} + m_{o2}\underbrace{(m_{f1}-1)}_{\leq 0} & &\blacksquare
\end{align}
On the other hand, Yager's rule adds the conflict $K$ to the fused mass, hence allowing to recuperate unknown mass (see \eqref{eq:yagers_rule}). 

Therefore, we propose to use Dempster's combination rule to fuse mass from the ray ISM since it ensures that, once the unknown mass has fallen below the threshold, it remains beneath it. This prohibits further changes through the deep ISM. On the other hand, we propose to use Yager's rule to fuse mass from deep ISMs, since it recuperates unknown mass in case of conflicting predictions. This allows us for falsely initialized cells to first move the falsely assigned mass back to the unknown state and from there shift it to the correct state without violating the lower bound $\underline{m_u}$.  

\subsection{Informational Interpretation of deep ISM Predictions}
\label{subsec:info_interpret_of_deep_ism}
In each step, we want each cell to be allocated with its true evidence mass $\hat{\vec{m}}$ given all the measurements $z$ observed so far. We train a neural network to predict $\tilde{\vec{m}}$, which approximates the true occupancy state at time step $N$. As input for the network, we transform the last $L$ measurements $z_{N-L},...,z_{N}$ into the frame at time step $N$ to obtain the accumulated input $z_{N-L:N}$. By doing so, we propose the following two approximations
\begin{align}
\label{eq:only_m_needed}
\vec{m}(z_{N-L},...,z_{N}) &\approx \vec{m}(z_0,...,z_N)\\
\label{eq:only_det_needed}
\vec{m}(z_{N-L:N}) &\approx \vec{m}(z_{N-L},...,z_{N})
\end{align}
First, we argue in \eqref{eq:only_m_needed} that all time steps earlier than $N-L$ do not contain additional relevant information for the gird cells being updated at step $N$. Secondly, by performing the approximation in \eqref{eq:only_det_needed}, we loose the information of where the detections have been recorded, thus, arguing that the detection positions contain the main amount of measurement information. Therefore, in theory, the following should hold
\begin{align}
	\hat{\vec{m}}(z_0,...,z_n) \approx \tilde{\vec{m}}(z_{N-L:N})
\end{align}
and we could update the map e.g. by replacing the cell state whenever the predictions unknown mass is smaller than the current cell state. With this approach, however, we would always overwrite the geometric ISM's predictions while the cell state is still below $\underline{m_u}$, thus prohibiting a proper information fusion. Moreover, the approach removes the filtering aspect of occupancy mapping which results in outliers with small unknown masses being directly allocated to cells, as can be seen in Fig. \ref{fig:occ_map_comparison}. This falsely assigned mass has then to be reallocated by the geometric method which in turn decreases the convergence speed. To reduce the influence of outliers, we propose an iterative update scheme in the following sections which both reduces falsely and increases correctly assigned mass, as shown in Tab. \ref{tab:scores}. 

\subsection{Identification of non-redundant Information}
Instead of using the neural network predictions to replace the cell states, we can also use the predictions to update it. However, there is a lot of redundant information between $\tilde{\vec{m}}(z_{N-L:N})$ and $\vec{m}(z_0,...,z_{N-1})$. Even, if we only use one measurement $z_N$ as network input, there would still be redundancy since the last couple of inputs already measured approximately the same environment. Using the accumulated input $z_{N-L:N}$, however, provides us with the possibility to asses the amount of non-redundant information $\text{I}(z_N|z_0,...,z_{N-1})$ of the latest measurement $z_N$, as follows
\begin{align}
	\text{I}(z_N|z_0,...,z_{N-1}) &= \text{I}(z_0,...,z_N) - \text{I}(z_0,...,z_{N-1})\\
	\text{I}(z_N|z_0,...,z_{N-1}) &\approx \text{I}(z_{N-L:N}) - \text{I}(z_0,...,z_{N-1})
\end{align} 
To account for the redundancy, we can use the identified amount of additional information to discount the network predictions with a discount factor $\gamma(\text{I}(z_N|z_0,...,z_{N-1}))$. The only remaining question is which quantity should be used as a measure of information $I(...)$ and how to define the discount factor based on that measure.

\subsection{Choice of Information Measure and Discount Factor}
\label{subsec:choice_of_info_measure_and_discount_fact}
For this work, we use the unknown mass as a measure of information content. The benefit over the usually used entropy \cite{su2018research} is that the case of total conflict, where all mass is equally distributed into free and occupied mass, is not viewed as total lack of information. Hence, we can still use the conflict case to represent dynamic objects with varying degrees of certainty based on the unknown mass. Therefore, we will write the non-redundant part of the prediction information as
\begin{align}
	\Delta m_u^{N} &= \tilde{m}_u^{N} - m_u^{N-1} 
\end{align}

With regards to the discount factor, we have two conditions. First, we want to reduce the redundant information in $\tilde{\vec{m}}$ so that the following holds
\begin{align}
	\gamma &=
	\begin{cases}
		0, \Delta m_u < 0\\
		0, \Delta m_u \rightarrow 0\\
		1, \Delta m_u \rightarrow 1
	\end{cases}
\end{align} 
To realize this relationship, we propose to use
\begin{align}
\label{eq:gamma_tanh}
	\gamma = \tanh(\alpha\max(0, \Delta m_u))
\end{align}
which allows to tune the convergence speed using the parameter $\alpha$. For our experiments, we choose $\alpha = 10$. If we first limit the network prediction's unknown mass to $\underline{m}_u$ using \eqref{eq:lower_un_mass_limit} and afterwards discount $\tilde{\vec{m}}$ using \eqref{eq:gamma_tanh}, we ensure that, after the cell state reached the unknown mass limit, it cannot be altered through $\tilde{\vec{m}}$. This, however, does not ensure that the cell state's unknown mass cannot be pushed below $\underline{m}_u$ before the limit is reached. Thus, as a second condition, we want to choose the discount factor $\gamma$ in a way that the fused unknown mass cannot fall beneath the given threshold $\underline{m}_u$, which can be written as follows
\begin{align}
	\vec{m}^{N} &= \vec{m}^{N-1} \oplus_Y (\gamma^{N} \otimes \tilde{\vec{m}}^{N})\\
	\label{eq:condition_on_gamma}
	\underline{m}_u &\leq m_{u}^{N} 
\end{align} 
Using Yager's update rule and \eqref{eq:discount_eq}, we can define a lower bound on $\gamma$
\begin{align}
\label{eq:condition_on_gamma_2}
	\underline{m}_u &\leq m_u^{N-1} (1 - \gamma^{N} + \gamma m_2^{N}) + K^{N} 
\end{align}
In case \eqref{eq:condition_on_gamma} holds, the following is also given
\begin{align}
	\underline{m}_u &\leq m_u^{N-1}\\
	\text{and for } \tilde{m}_u^{N} &= 1 \rightarrow m_{u}^{N} = m_u^{N-1} + K^{N} \geq \underline{m}_u
\end{align}
Under these conditions, we can rewrite \eqref{eq:condition_on_gamma_2} as follows
\begin{align}
	\gamma \leq \dfrac{m_{u}^{N-1} + K - \underline{m}_u}{m_{u}^{N-1}(1-\tilde{m}_{u}^{N})}
\end{align}
which provides an additional upper bound on $\gamma$. Eventually, the final discount factor is defined as
\begin{align}
\label{eq:gamma}
	\gamma^{N} &= \min\left( \dfrac{m_{u}^{N-1} + K - \underline{m}_u}{m_{u}^{N-1}(1-\tilde{m}_{u}^{N})}\text{, } \tanh(\alpha \max(0,m_{u}^{N-1} - \tilde{m}_{u}^{N})) \right)
\end{align} 
%
\section{EXPERIMENTAL SETUP}
For our experiments, we use the publicly available nuScenes dataset \cite{caesar2019nuscenes} from which we use the data of the $360^{\circ}$, roof mounted, 32 beam lidar, 5 radars mounted at the corners and one central at the front and dead reckoning lidar odometry. For all mapping, we first projected the inputs into a $512 \times 512$ bird's-eye-view grid corresponding to an area of $40 \times 40$ meters. Afterwards, one of the ISMs is applied on the detection image followed by a transformation, using the lidar odometry, back into the first measurements coordinate frame. For training and validation, we used the same data split as proposed by the nuScenes team, while removing scenes in which the ego vehicle does not move further than $1$ meter. This results in $133708$ training and $29846$ validation samples.
\subsection{Inverse Lidar Model}
\begin{figure}
	\begin{center}
		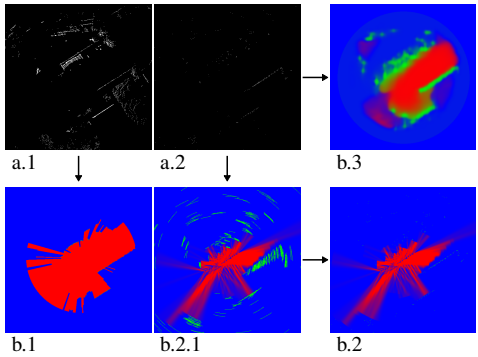
		\caption{\label{fig:comparison_isms} Illustration of the lidar detections a.1 and the corresponding ray ILM b.1, the radar detections a.2, the intermediate and final ray IRM b.2.1, b.2 and the deep IRM b.3  (best viewed in color with zoom)}
	\end{center}
\end{figure}

To compute the ray casting-based inverse lidar model, from now on referred to as ray ILM, several steps are performed. First, we remove all detections below $0.3m$ and above $3.0m$. The remaining points are then projected on the ground plane and discretized into the image grid. Eventually, rays are cast in all directions starting at the lidar coordinate frame's center with a resolution of $0.2^{\circ}$ up until a detection is hit or the maximal range ($15$ meters in our experiments) is reached. Finally, detection positions are assigned $m_o = 0.5$ and the ray points $m_f = 0.05$, as shown in Fig. \ref{fig:comparison_isms}.
\subsection{Inverse Radar Model}
\label{subsec:ray_ism}
For the ray casting-based inverse radar model (ray IRM), the last $N=10$ radar point clouds from each radar are projected into the current ego vehicle frame. Afterwards, cones with an opening angle of $2^{\circ}$ are cast from the respective temporal sensor positions towards the corresponding detections. As an intermediate step, the segment of each cone at max range is marked as occupied. Next, the same cones are cast again towards the detections, marking each pixel $m_f = 0.3$. This time, however, the cones are restricted by the occupied segments, as shown in Fig. \ref{fig:comparison_isms}, b.2.1. Eventually, the occupied segments are removed and the last $N$ detections are marked with $m_o = 0.5$ (Fig. \ref{fig:comparison_isms}, b.2).
\subsection{Deep Inverse Radar Model}
\label{subsec:evnet}
\begin{figure}
	\begin{center}
		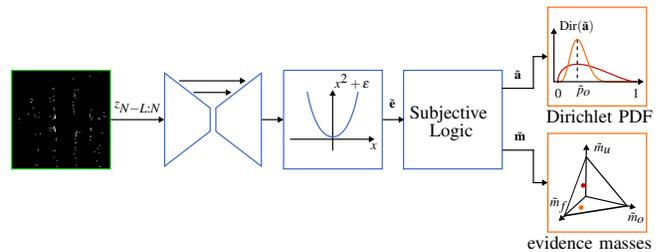
		\caption{\label{fig:evnet} Architecture of Ev-Net as proposed in \cite{bauer2019deep}}
	\end{center}
\end{figure}

As a deep inverse radar model (deep IRM), we train Ev-Net as described in \cite{bauer2019deep} to predict the ray ILM based on accumulated radar detection images. We adapt the architecture slightly to cope with the high resolution images by max pooling the inputs and bilinear upsampling the output. Moreover, we do not use occupancy map patches to train the model, since the results indicate only minor benefit. Ev-Net can be used to predict strictly positive evidences for free and occupied space $\tilde{\vec{e}} = \begin{bmatrix}\tilde{e}_f, \tilde{e}_o\end{bmatrix}^{\top}$ and use them to either obtain a probabilistic or evidential occupancy state prediction (see \eqref{eq:josang_evidence} - \eqref{eq:josang_evidential_view}).
The network, whose architecture is depicted in Fig. \ref{fig:evnet}, can be trained using the following loss
\begin{align}
\mathcal{L}_k = \sum_{k}(\hat{m}_k - \tilde{p}_k)^2 \cdot \dfrac{\tilde{p}_k(1-\tilde{p}_k)}{S + 1} &\text{, } &\mathcal{L}_u = (1-\tilde{m}_u)^2
\end{align}
To account for the class imbalance between the amount of free, occupied and unknown space, the expectation over each class is computed separately and afterwards accumulated to obtain the overall loss
\begin{align}
	\mathcal{L} &= \mathbb{E}[\mathcal{L}_f] + \mathbb{E}[\mathcal{L}_o] + \mathbb{E}[\mathcal{L}_u]
\end{align}
An exemplary prediction of the network is shown in Fig. \ref{fig:comparison_isms}, b.3. The network runs without further optimizations with over $10$ Hz on one core of an i7 8th gen intel CPU.
%
%
\section{EXPERIMENTAL RESULTS}
\begin{figure*}
	\begin{center}
		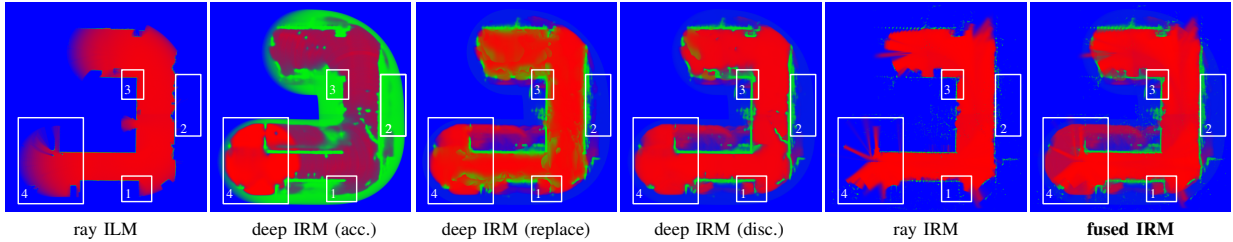
		\caption{\label{fig:occ_map_comparison}Qualitative comparison of occupancy mapping results for the investigated fusion approaches with their respective ISMs. Box 1) shows an exemplary area where the deep IRM (acc.) maps were correctly initialized in the beginning but later on falsely overwritten. Box 2) illustrates an area where deep IRM (acc.) accumulates small biases in occluded areas. Both of these problems are solved by our framework in fused IRM. Additionally, box 3) shows that the fused IRM's accuracy at object boundaries is similar to the ray IRM's as opposed to deep IRM (acc.). Eventually, box 4) marks an area which is already initialized by the deep IRM but only partially observed by the ray IRM.}
	\end{center}
\end{figure*}
In this section, we compare occupancy maps which are created by accumulating
\begin{enumerate}
	\item ray ILM with Dempster's rule as ground-truth reference
	\item deep IRM with Dempster's rule (deep IRM (acc.) )
	\item deep IRM with lower limit $\underline{m}_u$ and replacement of cells with lower unknown mass (deep IRM (replace))
	\item deep IRM with lower limit $\underline{m}_u$ and discounted redundant information using Yager's rule (deep IRM (disc.))
	\item ray IRM with Dempster's rule
	\item fusion of deep IRM (disc.) and ray IRM
\end{enumerate}  
\subsection{Interpretation of qualitative Results}
The exemplary occupancy mapping result using deep IRM (acc.) in Fig. \ref{fig:occ_map_comparison} reflects the problem that small biases accumulate over time in occluded areas (e.g. box 2)). In our case, the model has a small bias towards the occupied state which can be seen by the slightly green circle in Fig. \ref{fig:comparison_isms}, b.3. This problem is solved in our proposed framework by only accumulating the non-redundant part of the information, as can be seen by the mapping results for fused IRM.

Moreover, box 1) illustrates an area where correctly allocated areas close to data can be later on falsely overwritten by purely interpolation-based estimates. To prevent this behavior, we set a lower limit on the deep IRM's unknown mass and only allow the ray IRM to decrease the unknown mass further. Since we only allow the deep IRM to allocate non-redundant information, we therefor prohibit any contributions of the deep IRM once this lower unknown threshold is reached. Hence, areas cannot be falsely overwritten later on, as can be seen in the occupancy map of fused IRM box 1).

Box 3) shows that the accuracy at object boundaries for fused IRM is similar to ray IRM's as opposed to the results for deep IRM (acc.).  

Finally, box 4) shows an area which is already allocated for the fused IRM's occupancy map but only partially observed by the ray IRM's. This illustrates our frameworks capability to use the deep IRM as a prior to initialize areas based on data-driven interpolation. Additionally, the influence of the lower unknown limit on deep IRM can be seen by the lighter red in the regions solely allocated by the deep IRM. 

%
%
%
\subsection{Interpretation of quantitative Results}
The per class mean intersection over union scores in Tab. \ref{tab:scores} verify the problems when applying the deep IRM directly in the standard evidential occupancy mapping framework by iteratively fusing it into the map using Dempster's rule of combination. Additionally, the scores show that our proposed fusion scheme converges to the ray IRM's performance in the presence of sufficient measurements. 
%
\begin{table}
	\begin{center}
		\caption{\label{tab:scores}Per Class mean Intersection over Union Scores}
		\begin{tabular}{l|c|c|c}
			ISM used for Mapping & fr. mIoU & oc. mIoU & un. mIoU \\ 
			\hline 
			deep IRM (acc.)  &  61.78 & 2.07 & 11.05\\
			deep IRM (replace) &  69.63 & 3.43 & 45.96\\
			deep IRM (disc.) &  71.64 & 4.75 & 46.23\\
			ray IRM &  76.68 & 6.02 & 63.55\\
			\textbf{fused IRM} &  76.88 & 5.44 & 45.69
		\end{tabular}		
	\end{center}
\end{table}
\section{CONCLUSIONS}
We propose an integration procedure which only uses the non-redundant information of deep ISMs to update occupancy maps. The update is performed in a way that, when solely using the deep ISM, the map state's certainty cannot fall beneath a given threshold. This enables us to initialize the map in areas not observed by the geometric ISM and, thus, increase the overall perceptive field and convergence speed. Moreover, we are still able to distinguish cells solely allocated by the deep ISM and not yet verified using the geometric ISM by the tuneable certainty threshold. Additionally, in presence of enough measurements, the map converges to the geometric ISM which results in sharper borders between free and occupied space.

\addtolength{\textheight}{-12cm}   

%
%

\bibliographystyle{IEEEtran}
\bibliography{IEEEabrv,bib/bib}
\end{document}